\documentclass[conference,letterpaper]{IEEEtran}

\usepackage[utf8]{inputenc} 
\usepackage[T1]{fontenc}
\usepackage{url}
\usepackage{ifthen}
\usepackage{cite}
\usepackage[cmex10]{amsmath} 
\usepackage{epsfig,rotating,setspace,latexsym,epsf,amssymb,bm}
\usepackage{amsthm}
\usepackage{bbm}
\usepackage{xcolor}
\usepackage{cite}
\usepackage{multicol}
\usepackage{hyperref}

\usepackage{algorithmic}
\usepackage{algorithm}
\usepackage{caption}
\newtheorem{definition}{Definition}
\newtheorem{theorem}{Theorem}

\newtheorem{remark}{Remark}

\usepackage[compact]{titlesec}         
\titlespacing{\section}{2pt}{2pt}{2pt} 
\AtBeginDocument{
  \setlength\abovedisplayskip{2pt}
  \setlength\belowdisplayskip{2pt}}

\begin{document}
\title{Intrinsic Fairness-Accuracy Tradeoffs under Equalized Odds}

\author{\IEEEauthorblockA{Meiyu Zhong ~~ Ravi Tandon}
\IEEEauthorblockA{Department of Electrical and Computer Engineering \\
University of Arizona, Tucson, USA \\
E-mail: \textit{\{meiyuzhong, tandonr\}}@arizona.edu
}}

\maketitle

\begin{abstract}
With the growing adoption of machine learning (ML) systems in areas like law enforcement, criminal justice, finance, hiring, and admissions, it is increasingly critical to guarantee the fairness of decisions assisted by ML. In this paper, we study the tradeoff between fairness and accuracy under the statistical notion of equalized odds. We present a new upper bound on the accuracy (that holds for \textit{any} classifier), as a function of the fairness budget. In addition, our bounds also exhibit dependence on the underlying statistics of the data, labels and the sensitive group attributes. We validate our theoretical upper bounds through empirical analysis on three real-world datasets: COMPAS, Adult, and Law School. Specifically, we compare our upper bound to the tradeoffs that are achieved by various existing fair classifiers in the literature. 
   Our results show that achieving high accuracy subject to a low-bias could be fundamentally limited based on the statistical disparity across the groups.   
\end{abstract}
\section{Introduction}
Machine learning-based solutions are increasingly being implemented across various sectors, including criminal justice, law enforcement, hiring, and admissions. These systems have demonstrated remarkable predictive capabilities. However, recent studies \cite{mehrabi2021survey,zemel2013learning, zafar2017fairness} indicate a significant downside to data-driven approaches: bias in decision-making. To address this problem, there is a vast amount of research focused on various concepts of fairness \cite{dwork2012fairness,zafar2017fairness1,NIPS2017_a486cd07}, which mainly falls into three categories: (1) Group Fairness \cite{barocas2016big,feldman2015certifying,hardt2016equality} which requires that the subjects in the subgroups have an equal probability of being assigned to the same predicted class. (2) Individual Fairness \cite{dwork2012fairness,NIPS2017_a486cd07,yurochkin2019training} which requires that \textit{similar individuals} (measured by a domain specific similarity metric) should be treated similarly. (3) Causality-based Fairness \cite{kilbertus2017avoiding,NIPS2017_a486cd07}: which uses causality-based tools to design notions of fairness.


In this paper, we focus on group fairness (also known as statistical fairness). There are three types of statistical fairness notions which have been widely studied: demographic parity (DP) \cite{zafar2017fairness,dwork2012fairness}; equalized odds (EO) \cite{hardt2016equality,zafar2017fairness1} and predictive rate parity (PP) \cite{zafar2017fairness1}. DP requires that the classifier's decision be independent of the sensitive group attribute. The notion of DP however suffers from two drawbacks: first, when the sensitive group attribute is correlated with the class labels, this may rule out the perfect predictor (and hurt accuracy) \cite{barocas2023fairness}; second, the fairness notion of DP does not take into account the true label into account; if the distribution of data across subgroups is uneven, then enforcing DP may be unfair to those individuals which were worthy of a positive outcome.  To avoid such disparity, the notion of Equalized Odds (EO) requires the classifier's prediction should be independent of sensitive attributes given the true class label. In addition, Predictive Rate Parity (PP) is defined as the condition where different groups have equal predictive values, meaning that the probability of a true positive (or negative) result is consistent across all groups. 

The techniques for learning a fair classifier subject to one of the group fairness notions can be mainly divided into three categories: (1) Pre-processing methods \cite{zemel2013learning}, which focus on mitigating bias by altering the training data (e.g., by creating a more balanced or fair dataset) before it is utilized in the training phase. (2) In-processing methods \cite{zafar2017fairness,cho2020fair1}, which involve integrating fairness constraints directly into the model training process (for instance, via explicit fairness aware regularization). (3) Post-processing methods \cite{pleiss2017fairness}, which entail adjusting the model's parameters after training. This technique involves fine-tuning the trained model to rectify any unfair biases that may have been introduced during the training process. However, experimental results \cite{zhong2023learning, zafar2017fairness,zafar2017fairness1,hardt2016equality} have demonstrated, and recent theoretical evidence \cite{zhao2022inherent,zeng2022bayes} has further confirmed, that there is often a drop in accuracy when enforcing fairness constraints, when compared to unconstrained training. 

\noindent \textit{Overview of recent work on Fairness-Accuracy Tradeoffs}: The above observations motivate a theoretical treatment of the tradeoff between fairness and accuracy. Recent works \cite{zhao2022inherent,zeng2022bayes,menon2018cost} have studied this tradeoff for the case of demographic parity (DP), and have obtained bounds which quantify the minimal drop in accuracy as a function of the fairness budget. 
Additionally, Dutta et al. \cite{dutta2020there} employ mismatched hypothesis testing to demonstrate that there exist distributions for which there is no trade-off between fairness and accuracy. However, they also show that these trade-offs do exist in real-world datasets. 
Wang et al. \cite{wang2023aleatoric} suggest that randomized prediction methods might more consistently meet equalized odds in classification (also see  \cite{tang2022attainability, kim2020fact, hamman2023demystifying, sabato2020bounding}). Another line of works \cite{zeng2022bayes,chzhen2019leveraging} study fair Bayes-optimal classifiers subject to equal opportunity (which is a related, but weaker notion than EO); the idea herein is to change the Bayes-optimal classifier by designing subgroup-specific decision thresholds to satisfy the fairness constraint. Another research direction, as explored in  \cite{zhao2019conditional}, concentrates on  determining whether it's possible to establish a bound on the accuracy of a given classifier relative to its allocated fairness budget. However, these bounds are intrinsically linked to the characteristics of the classifier itself and do not directly give an insight to the fundamental fairness-accuracy tradeoff.


\noindent\textbf{\textit{Main Contributions.}} 
In this paper, we focus on the problem of binary ($0/1$) classification subject to equalized odds (EO) fairness constraints and present a new upper bound on the accuracy as a function of the fairness budget. Our bounds are \textit{classifier-independent} (i.e., they must hold for any classifier) and are determined by the underlying statistics of the data, labels, and sensitive groups. Our primary technique for deriving these bounds involves adapting Le Cam's bound \cite{le1956asymptotic,yu1997assouad}, which is traditionally used for binary classification problems; encompassing the Equalized Odds (EO) fairness constraints. The original Le Cam's bound is based on the total variation distance between two class distributions (i.e., \(d_{TV}(P_0, P_1)\)). Our EO constrained bound depends on the total variation distances within sensitive subgroups (i.e., \(d_{TV}(P_0^a, P_1^a), d_{TV}(P_0^b, P_1^b)\), where $a, b$ denotes the two subgroups), as well as the relative proportions of the subgroups. The extent of statistical discrepancy across the subgroups plays a critical role in influencing the behavior of our bound. As the subgroup discrepancy increases, our bound becomes tighter compared to the classical Le Cam's bound.

We also present experimental results using three real-world datasets: COMPAS, Adult, and the Law School dataset. We estimate our bounds for these datasets by employing an estimator for the total variation distance. Additionally, we compare our upper bounds with the trade-offs between fairness and accuracy achieved by various fair classifiers on these datasets.


\section{Preliminaries and Problem Statement}
We consider a supervised classification problem, where we are given a dataset of $n$ users:
    ${\{x_i,\;y_i,\;z_i\}}_{i=1}^n$,
where $x_i \in \mathcal{X} \subseteq \mathbb{R}^d$ denotes the set of features of $i$th training sample;  $y_i \in \mathcal{Y} = \{0,1\}$ represents the corresponding binary class label; $z_i \in \mathcal{Z} = \{a,b\}$ denotes the set of binary sensitive attributes of $i$th training sample, which depends on the dataset and underlying context. For instance, if we consider gender as a sensitive attribute (or subgroup), we could designate $a$ to represent females and $b$ to represent males. Consider a classifier $f$ which maps from input data space to output labels: $f: \mathcal{X} \rightarrow \mathcal{Y}$, i.e., $f(x) \in \{0,1\}$ denotes the classifier's decision for an input $x$.  

\textit{Notation:} For the scope of this paper, we use capital letters to represent the random variables/vectors ($X,Y,Z$) and use lowercase letters ($x,y,z$) to denote a specific realization. We next define the following quantities which will appear in our results:
\begin{itemize}
\item $\alpha =  \mathbb{P}(Y = 1)$ (probability of the label $Y=1$).
\item $\beta = \mathbb{P}(Z = a)$ (probability of  sensitive attribute $Z=a$). 
\item $\phi(x) = \mathbb{P}(Y=1|x)$ (posterior probability of class $1$)
\item $P(x)$ represents the unconditional PDF of $X$.
\item $P_{0}(x)$ (respectively, $P_{1}(x)$) represents the conditional PDF of $X|Y=0$ (respectively, $X|Y=1$).
\item $P^{a}_{0}(x)$ (respectively, $P^{b}_{0}(x)$) represents the conditional PDF of $X|(Y=0,Z=a)$ (respectively, $X|(Y=0,Z=b)$). Similarly, $P^{a}_{1}(x)$ (respectively, $P^{b}_{1}(x)$) represents the conditional PDF of $X|(Y=1,Z=a)$ (respectively, $X|(Y=1,Z=b)$).
\end{itemize}

\noindent\textbf{Problem Statement} From the scope of this paper, our goal is to explore tradeoff between fairness and accuracy for any classifier. We mainly focus on the group fairness notion: equalized odds, which is defined as follows.
\begin{definition}\label{def:eq_odds}(Approximate Equalized Odds)
    A binary classifier $f$ satisfies $\epsilon_{EO}$-Equalized Odds (EO) if $\Delta_{EO}\leq \epsilon_{EO}$, where, for $y \in \{0,1\}$, $\Delta_{EO}:= $
\begin{align}
       \hspace{-2pt}\underset{y}{\max}|\mathbb{P}(f(X) = 1 | Z\hspace{-2pt}=\hspace{-2pt}a, Y\hspace{-2pt}=\hspace{-2pt}y)\hspace{-2pt} -\hspace{-2pt}\mathbb{P}( f(X) = 1| Z\hspace{-2pt}=\hspace{-2pt}b, Y\hspace{-2pt}=\hspace{-2pt}y)|.\nonumber
\end{align}   
\end{definition}    
    When $\epsilon_{EO} =0 $, the input $X$'s prediction $f(X)$ is conditionally independent of its sensitive attribute $Z$ given the label $Y$ (equivalent to the  Markov chain $f(X)\rightarrow Y \rightarrow Z$) which corresponds to the notion of \textit{perfect} equalized odds \cite{barocas2016big,zafar2017fairness1}. Following prior works, 
\cite{cho2020fair,cho2020fair1,zafar2017fairness1}, we focus on the approximate EO setting, $\Delta_{EO} \leq \epsilon_{EO}$, and $\epsilon_{EO}$ refers to the fairness budget. 



\begin{definition} (Total Variation (TV) Distance) 
Consider P and Q as two probability distributions over a common probability space $\Omega$. Then, the total variation distance between them, denoted $d_{TV}(P, Q)$, is defined as follows
    \begin{align}
        d_{TV}(P, Q) = \underset{A \subseteq \Omega}{\text{sup}} |P(A)-Q(A)|= \frac{1}{2} \int_x |P(x) - Q(x)| dx \nonumber
    \end{align}
\end{definition}
\section{Main Results and Discussion}

In this section, we present our main results on the fundamental trade-off between fairness constraints (subject to EO) and accuracy for binary classification. Our results are organized as follows. First, we present an upper bound on accuracy based on Le Cam's method without any fairness constraint. This bound is attainable by the Bayes optimal classifier for three cases: when the class distribution is either balanced $\alpha = 0.5$ or extremely unbalanced $\alpha = 0/1$. We then present the main result of this paper: a new \textit{classifier-independent} upper bound on accuracy as a function of the EO budget $(\epsilon_{EO})$, taking into account the underlying statistics of the data, labels, and the proportions of subgroups. 

We first present an upper bound on the accuracy for any arbitrary binary classifier, which is given as follows:
\begin{theorem}\label{The:generilized_le_cam}(Unconstrained Upper Bound on Accuracy)
    For any binary classifier $f$, its accuracy $Acc(f)$ satisfies $\text{Acc}(f) \leq \overline{\text{Acc}}$, where $\overline{\text{Acc}}$ is given as follows:
    \begin{align}
        \overline{\text{Acc}} = \max(1-\alpha, \alpha)+ \min(1-\alpha, \alpha)\cdot d_{TV}(P_1,P_0).\label{eq:unconstrained_acc}
    \end{align}
    Furthermore, the upper bound can be attained by the Bayes optimal classifier when $\alpha = 0, 0.5$ or $1$.
\end{theorem}
\begin{figure*}
    \centering
    \includegraphics[scale = 0.25]{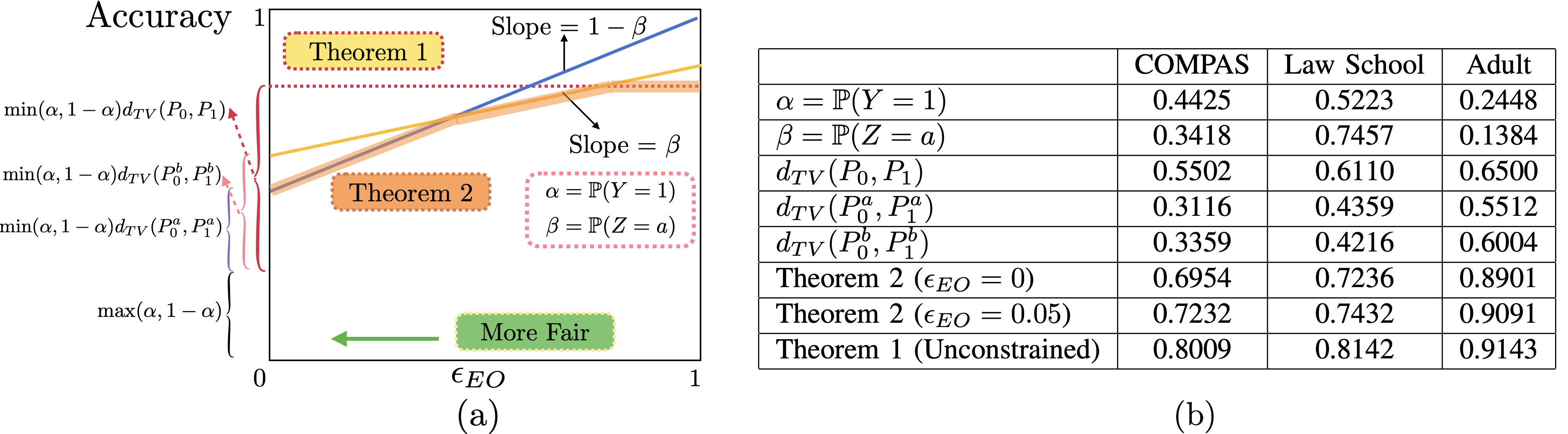}
    \caption{(a) Illustration of the relationship between Theorem \ref{The:generilized_le_cam} and Theorem \ref{the:tradeoff_acc_fair}, where the red dotted line represents the bound established in Theorem \ref{The:generilized_le_cam}, while the orange fluorescent line depicts the minimum of the two functions $T_1$ and $T_2$ in Theorem \ref{the:tradeoff_acc_fair}, where $T_1$ and $T_2$ have slopes of $1-\beta$ and $\beta$, respectively. (b) Dataset-related parameters and upper-bound-related parameters in the real world datasets: COMPAS, Adult and Law School dataset.}
    \label{fig:bound_overall}

\end{figure*}
We can draw the following insights from the above result: for a fixed value of $\alpha$, accuracy is directly proportional to $d_{TV}(P_1, P_0)$. On the other hand, for a fixed value of $d_{TV}(P_1, P_0)$, the lowest accuracy occurs when $\alpha=0.5$, i.e., when both the classes $0/1$ appear in a balanced manner. We next present the proof of the above result, which will allow us to contrast the corresponding bounds we obtain subject to equalized odds fairness constraints. 
\vspace{-10pt}
\begin{proof}
We expand the expression for accuracy of a binary classifier using total probability theorem as follows:
    \begin{align}
        &\text{Acc}(f) = E_{(X,Y)}[\mathbbm{1}(f(X) = Y)] = \mathbb{P}(f(X) = Y)\nonumber\\
        &= \mathbb{P}(Y =  1, f(X) = 1) + \mathbb{P}(Y =  0, f(X) = 0)\nonumber\\
        &= \alpha \mathbb{P}(f(X) = 1 | Y =  1)\hspace{-2pt}+\hspace{-2pt}(1-\alpha) \mathbb{P}(f(X) = 0 | Y =  0)
    \end{align}
 We can re-write the above expression for $Acc(f)$ in two different ways as follows, where $\text{Acc}(f)$
 \begin{align}
     &\overset{(a)}{=}  \alpha\mathbb{P}(f(X) = 1| Y =  1)  \hspace{-2pt}+ \hspace{-2pt} (1-\alpha)(1 \hspace{-2pt}- \hspace{-2pt}\mathbb{P}(f(X) = 1| Y =  0))\nonumber \\
     &\overset{(b)}{=} \alpha (1\hspace{-2pt}- \hspace{-2pt}\mathbb{P}(f(X) = 0| Y =  1))\hspace{-2pt} + \hspace{-2pt}(1-\alpha)\mathbb{P}(f(X) = 0| Y =  0) \nonumber 
 \end{align}
We now consider two scenarios based on the value of $\alpha$. In the first case, when $\alpha \leq 0.5$, we use the expression in $(a)$ and upper bound it as follows:
    \begin{align}
        &\hspace{-2pt}\text{Acc}(f)\hspace{-2pt}\leq \hspace{-2pt}1\hspace{-2pt}-\hspace{-2pt}\alpha \hspace{-2pt}+\hspace{-2pt} \alpha (\mathbb{P}(f(X) = 1| Y =  1) \hspace{-2pt}- \hspace{-2pt}\mathbb{P}(f(X) = 1| Y =  0)) \nonumber\\
        &\hspace{-2pt}\leq 1\hspace{-2pt}-\hspace{-2pt}\alpha \hspace{-2pt}+\hspace{-2pt}\underset{f}{\text{sup}}~\alpha  (\mathbb{P}(f(X) = 1| Y =  1) \hspace{-2pt}- \hspace{-2pt}\mathbb{P}(f(X) = 1| Y =  0)) \nonumber\\
        &\hspace{-2pt}\leq 1\hspace{-2pt}-\hspace{-2pt}\alpha \hspace{-2pt}+\hspace{-2pt}\alpha~\underset{f}{\text{sup}}~  |\mathbb{P}(f(X) = 1| Y =  1) \hspace{-2pt}- \hspace{-2pt}\mathbb{P}(f(X) = 1| Y =  0)|\nonumber\\
        &\hspace{-2pt}=1\hspace{-2pt}-\hspace{-2pt}\alpha \hspace{-2pt}+\hspace{-2pt}\alpha \cdot d_{TV}(P_0,P_1). \label{eq:gener_boun1}       
    \end{align}
Similarly, for the case when $\alpha>0.5$, we use the expression in $(b)$ and bound it in a similar manner to arrive at
    \begin{align}
        &Acc(f)\leq \hspace{-2pt}\alpha \hspace{-2pt}+\hspace{-2pt} (1-\alpha) \cdot d_{TV}(P_1,P_0) \label{eq:gener_boun2}     
    \end{align}
Combining \eqref{eq:gener_boun1} and \eqref{eq:gener_boun2}, we obtain the upper bound stated in the Theorem \ref{The:generilized_le_cam}:
    \begin{align}
        Acc(f)\leq  \max(\alpha,1-\alpha) + \min(\alpha,1-\alpha)\cdot d_{TV}(P_1,P_{0}). \nonumber
    \end{align}
 Let us now consider the Bayes optimal classifier:
\begin{align}
    f_{\text{Bayes}}(x) = \begin{cases}
        1, & \text{if} ~~\phi(x) \geq 1 - \phi(x)\\
        0,& \text{otherwise}. \label{eq: bayes_optimal}
    \end{cases}    
\end{align} 
where $\phi(x)= \mathbb{P}(Y=1|X=x)$.  The accuracy of Bayes optimal classifier can be readily computed as: 
\begin{align}
    &Acc(f_{\text{Bayes}})= \mathbbm{E}_{X}\left[\text{max}(\phi(X), 1-\phi(X))\right]  \\
    & =  \frac{1}{2} + \frac{1}{2} \int_{x} |\alpha \mathbb{P}(x|y =1) - (1-\alpha)\mathbb{P}(x|y =0) | dx \label{eq:bayes_acc}
\end{align}
It is straightforward to verify that the Bayes' classifier accuracy matches with our upper bound when $\alpha = 0.5, 0~\text{or}~1$. This completes the proof of Theorem \ref{The:generilized_le_cam}. 
\vspace{-8pt}
\end{proof}
We next present our main result, 
which is an upper bound on accuracy as a function of the fairness budget $\epsilon_{EO}$. 
\vspace{-5pt}
\begin{theorem}\label{the:tradeoff_acc_fair}
    For any binary classifier which satisfies equalized odds ($ \Delta_{EO}\leq \epsilon_{EO} $), its accuracy $\text{Acc}(f,\epsilon_{EO})$ satisfies $\text{Acc}(f,\epsilon_{EO}) \leq \overline{\text{Acc}}(\epsilon_{EO})$, where $\overline{\text{Acc}}(\epsilon_{EO})$ is given as follows:
\begin{align}
    \overline{\text{Acc}}(\epsilon_{EO})= \max(1-\alpha, \alpha) + \min(T_1, T_2),
\vspace{-8pt}
\end{align}
where 
\vspace{-5pt}
\begin{align}
    T_1&\triangleq \min(1-\alpha, \alpha) \cdot d_{TV}(P^{b}_1, P^{b}_{0})+\beta \epsilon_{EO}\\
    T_2&\triangleq \min(1-\alpha, \alpha) \cdot d_{TV}(P^{a}_{1}, P^{a}_{0}) + (1-\beta) \epsilon_{EO}.
\end{align}

\end{theorem}
We next present a sequence of remarks which give an operational interpretation of the above bound. 

\begin{remark}
    We note that the upper bound is a piece-wise linear function of the fairness budget $\epsilon_{EO}$, and obtained as a minimum over two expressions related to $T_1, T_2$ as shown in Fig. \ref{fig:bound_overall}(a). The expressions in $T_1$ depend upon the $d_{TV}(P^{b}_1, P^{b}_{0})$, which reflect the difference in the class conditional probabilities for subgroup $b$. Similarly, the bound in $T_2$ depends on $d_{TV}(P^{a}_1, P^{a}_{0})$, which reflects the difference in the class conditional probabilities for subgroup $a$. When $\epsilon_{EO}=0$, i.e., perfect EO fairness constraint, the upper bound shows that the accuracy of any classifier will always be limited by the minimum of $d_{TV}(P^{b}_1, P^{b}_{0})$ and $d_{TV}(P^{a}_1, P^{a}_{0})$, i.e., the subgroup with the worst classification accuracy. 
\end{remark}

\begin{remark}
    For the case when $\epsilon_{EO}>0$, i.e., approximate EO fairness constraint, the bound is given by the minimum over two linear functions of $\epsilon_{EO}$. The slopes of these two lines are given by $\beta = P(Z=a)$, and $1-\beta = P(Z=b)$, i.e., relative proportion/size of the two subgroups shown in Fig. \ref{fig:bound_overall} (a).  The interplay between $(\beta, 1-\beta)$ and the subgroup wise statistical distances dictate the overall behaviour of the upper bound. For instance, suppose subgroup $a$ is a minority group with $\beta << 1-\beta$, then there always exists a threshold $\epsilon_{EO}$, for which the classification accuracy will be  dictated by the statistical distance corresponding to the majority subgroup $b$. 
\end{remark}

 \begin{remark}
In the next sub-section, we discuss a methodology to estimate the upper bounds for real-world datasets by leveraging tools for statistical estimation of $f$-divergence (e.g., TV distance). 
In Fig. \ref{fig:bound_overall}(b), we show the corresponding experimental results we obtained on three datasets (COMPAS, Law-School admissions and Adult Income prediction). Specifically, for each of these datasets, we estimate the corresponding values of $\beta, \alpha, d_{TV}(P_0,P_1), d_{TV}(P^{a}_0,P^{a}_1)~\text{and}~d_{TV}(P^{b}_0,P^{b}_1)$ from the datasets and then show the bounds obtained from the two Theorems. 
We show the corresponding bounds for $\epsilon_{EO}=0$ and for $\epsilon_{EO}=0.05$ (more experiments are presented in the next Section). Specifically, we can observe the estimated bound from Theorem \ref{The:generilized_le_cam} (corresponding to unconstrained classifiers) is always larger than those provided by Theorem \ref{the:tradeoff_acc_fair} (the upper bound on accuracy with EO constraints), as well as the interplay between the statistical properties, namely $\beta, 1-\beta$ and the statistical disparity across the two subgroups. 
\end{remark}
\noindent We next present the proof of Theorem \ref{the:tradeoff_acc_fair}. 
\begin{proof}
    \begin{figure*}[t]
    \centering
    \includegraphics[scale=0.27]{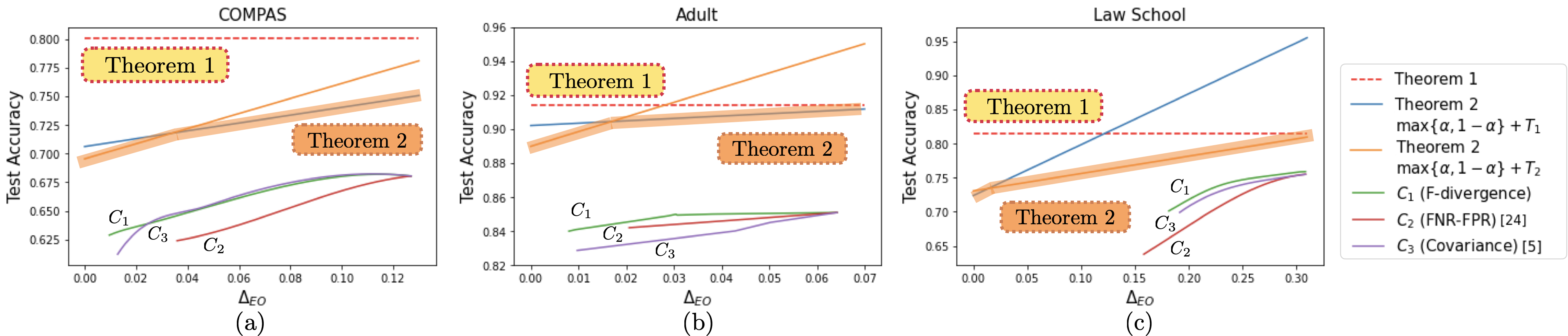}
    \caption{Comparison of the upper bound of the test accuracy in real world datasets: (a) COMPAS dataset, (b) Adult dataset, (c) Law School dataset under three fair classifiers ($C_1$ \cite{zhong2023learning}, $C_2$ \cite{bechavod2017learning} and $C_3$ \cite{zafar2017fairness1}). We can observe that Theorem \ref{the:tradeoff_acc_fair} is consistently tighter than the unconstrained upper bound (Theorem \ref{The:generilized_le_cam}), and Theorem \ref{the:tradeoff_acc_fair} provides the tightly upper bound on the tradeoffs achieved by three classifiers for three real world datasets.}
    \vspace{-5pt}
    \label{fig:upperbound}
\end{figure*}
    To simplify the notation used in the proof, we denote $\eta = \mathbb{P}(Z=a|Y=1)$ and denote $\gamma = \mathbb{P}(Z=a|Y=0)$. We start by the definition of the total accuracy of the binary classifier using total probability theorem  as follows:
    \begin{align}
        & \text{Acc}(f) = E_{(X,Y)}[\mathbbm{1}(f(X) = Y)] = \mathbb{P}(f(X) = Y)\nonumber\\
        & \hspace{-4pt}=  \hspace{-2pt}\alpha \mathbb{P}(f(X)=1| Y=1)\hspace{-2pt}+ \hspace{-2pt}(1-\alpha) \mathbb{P}(f(X) = 0| Y = 0) \label{eq:separate}\\
        & \hspace{-6pt}\overset{(a)}{=} \hspace{-2pt}\alpha \mathbb{P}(f(X)=1| Y=1)\hspace{-2pt} + \hspace{-2pt}(1 -\alpha ) (1-\mathbb{P}(f(X) = 1| Y = 0)) \nonumber\\
        &\hspace{-6pt}\overset{(b)}{=} \alpha (1\hspace{-2pt}- \hspace{-2pt}\mathbb{P}(f(X) = 0| Y =  1))\hspace{-2pt} + \hspace{-2pt}(1-\alpha)\mathbb{P}(f(X) = 0| Y =  0) \nonumber 
    \end{align}
    Then we further apply total probability theorem on $\mathbb{P}(f(X)=1| Y=1)$ with respect to $Z$ in $(a)$. Therefore, we can express $\mathbb{P}(f(X)=1| Y=1)$ as follows: 
    \begin{align}
         \eta \mathbb{P}(f(X)\hspace{-2pt}=\hspace{-2pt}1|Y\hspace{-2pt}=\hspace{-2pt}1,Z\hspace{-2pt}=\hspace{-2pt}a)\hspace{-2pt}+\hspace{-2pt}(1-\eta) \mathbb{P}(f(X)\hspace{-2pt}=\hspace{-2pt}1|Y\hspace{-2pt}=\hspace{-2pt}1,Z\hspace{-2pt}=\hspace{-2pt}b) \nonumber
    \end{align}
    Similarly, we can also take the advantage of total probability theorem on $\mathbb{P}(f(X) = 1| Y = 0)$ with respect to $Z$ in $(a)$, where $\mathbb{P}(f(X) = 1| Y = 0)$ can be written as:
    \begin{align}
        \gamma \mathbb{P}(f(X)\hspace{-2pt}=\hspace{-2pt}1|Y\hspace{-2pt}=\hspace{-2pt}0,Z\hspace{-2pt}=\hspace{-2pt}a)\hspace{-2pt}+\hspace{-2pt}(1-\gamma) \mathbb{P}(f(X)\hspace{-2pt}=\hspace{-2pt}1|Y\hspace{-2pt}=\hspace{-2pt}0,Z\hspace{-2pt}=\hspace{-2pt}b) \nonumber
    \end{align}
    Since the classifier satisfies the fairness constraints $\Delta_{EO} \leq \epsilon_{EO}$, we have:
    \begin{align}
        &|\mathbb{P}(f(X)\hspace{-2pt}=\hspace{-2pt}1|Y\hspace{-2pt}=\hspace{-2pt}1,Z\hspace{-2pt}=\hspace{-2pt}a) \hspace{-2pt}-\hspace{-2pt} \mathbb{P}(f(X)\hspace{-2pt}=\hspace{-2pt}1|Y\hspace{-2pt}=\hspace{-2pt}1,Z\hspace{-2pt}=\hspace{-2pt}b)|\leq \epsilon_{EO} \nonumber\\
        &|\mathbb{P}(f(X)\hspace{-2pt}=\hspace{-2pt}1|Y\hspace{-2pt}=\hspace{-2pt}0,Z\hspace{-2pt}=\hspace{-2pt}a)\hspace{-2pt} - \hspace{-2pt}\mathbb{P}(f(X)\hspace{-2pt}=\hspace{-2pt}1|Y\hspace{-2pt}=\hspace{-2pt}0,Z\hspace{-2pt}=\hspace{-2pt}b)|\leq \epsilon_{EO}. \nonumber
    \end{align}
    By incorporating the above inequalities, we can bound $\mathbb{P}(f(X)=1| Y=1)$ and $\mathbb{P}(f(X) = 1| Y = 0)$ as follows:
    \begin{align}
        &\hspace{-4pt}\mathbb{P}(f(X)\hspace{-2pt}=\hspace{-2pt}1| Y\hspace{-2pt}=\hspace{-2pt}1) \hspace{-2pt}\leq\hspace{-2pt}\eta\epsilon_{EO} + \mathbb{P}(f(X)\hspace{-2pt}=\hspace{-2pt}1|Y\hspace{-2pt}=\hspace{-2pt}1,Z\hspace{-2pt}=\hspace{-2pt}b) \label{eq:boundfxy1}\\
        &\hspace{-4pt}\mathbb{P}(f(X) \hspace{-2pt}=\hspace{-2pt} 1| Y\hspace{-2pt} = \hspace{-2pt}0) \hspace{-2pt}\geq \hspace{-2pt}-\gamma\epsilon_{EO}\hspace{-2pt} +\hspace{-2pt} \mathbb{P}(f(X)\hspace{-2pt}=\hspace{-2pt}1|Y\hspace{-2pt}=\hspace{-2pt}0,Z\hspace{-2pt}=\hspace{-2pt}b).\label{eq:boundfxy0}
    \end{align}
    By plugging \eqref{eq:boundfxy1} and \eqref{eq:boundfxy0} into (a), the total accuracy can be bounded by:
    \begin{align}
        \text{Acc}(f) &\leq (1-\alpha) \hspace{0pt} +\hspace{0pt}\alpha\mathbb{P}(f(X)\hspace{-2pt}=\hspace{-2pt}1|Y\hspace{-2pt}=\hspace{-2pt}1,Z\hspace{-2pt}=\hspace{-2pt}b)\hspace{-2pt}\nonumber\\
        &\hspace{5pt}-(1\hspace{-2pt}-\hspace{-2pt}\alpha)\mathbb{P}(f(X)\hspace{-2pt}=\hspace{-2pt}1|Y\hspace{-2pt}=\hspace{-2pt}0,Z\hspace{-2pt}=\hspace{-2pt}b)+ \beta \epsilon_{EO}, \label{eq:eo17}
    \end{align}
    where $\beta = \alpha\eta+(1-\alpha)\gamma = P(Z=a)$. Similar to the proof of Theorem \ref{The:generilized_le_cam}, we next consider two cases with respect to $\alpha$. If $\alpha \leq 0.5$, then \eqref{eq:eo17} can be further upper bounded by taking a supremum over $f$. We thus arrive at the following: 
    \begin{align}
        &\text{Acc}(f) 
         \leq (1-\alpha) \hspace{-2pt} +\hspace{-2pt} \beta \epsilon_{EO} \nonumber+\\&\alpha\cdot\underset{f}{\text{sup}}~|\mathbb{P}(f(X)\hspace{-2pt}=\hspace{-2pt}1|Y\hspace{-2pt}=\hspace{-2pt}1,Z\hspace{-2pt}=\hspace{-2pt}b)\hspace{-2pt}-\hspace{-2pt}\mathbb{P}(f(X)\hspace{-2pt}=\hspace{-2pt}1|Y\hspace{-2pt}=\hspace{-2pt}0,Z\hspace{-2pt}=\hspace{-2pt}b)|\nonumber\\
        &= (1-\alpha) + \alpha d_{TV}(P_1^b, P_0^b) + \beta\epsilon_{EO}. \label{eq: eouppbound1}
    \end{align}
    When $\alpha > 0.5$, we focus on (b), where we derive the upper bound subject to accuracy following the same steps above as (a).
    we arrive at the upper bound:
    \begin{align}
        \text{Acc}(f) \leq \alpha + (1-\alpha) \cdot d_{TV}(P_1^b, P_0^b) + \beta\epsilon_{EO}. \label{eq: eouppbound2}
    \end{align}
    To this end, combining two inequalities \eqref{eq: eouppbound1} and \eqref{eq: eouppbound2}, we have the upper bound of accuracy as a function of $\epsilon_{EO}$ with respect to the subgroup $Z=b$ and the proportion of another subgroup $Z=a$, which can be compactly written as:
    \begin{align}
        \max(\alpha,1-\alpha ) + \min(\alpha,1-\alpha)\cdot d_{TV}(P_1^b, P_0^b) + \beta\epsilon_{EO} \nonumber
    \end{align}
    Note that we can also bound $\mathbb{P}(f(X)=1| Y=1), ~\mathbb{P}(f(X) = 1| Y = 0)$ and $\mathbb{P}(f(X) = 0| Y =  1), \mathbb{P}(f(X) = 0| Y =  0)$ by another subgroup $Z=a$ in \eqref{eq:boundfxy1} and \eqref{eq:boundfxy0}. By doing so, we arrive at another upper bound of accuracy as a function of $\epsilon_{EO}$ with respect to the subgroup $Z = a$ and the proportion of another subgroup $Z = b$, which can be expressed as:
    \begin{align}
        \max(\alpha,1-\alpha) \hspace{-2pt}+ \hspace{-2pt}\min(\alpha, 1-\alpha) \cdot d_{TV}(P_{1}^a,P_{0}^a) + (1-\beta) \epsilon_{EO} \nonumber
    \end{align}
    Combining the two upper bounds, we arrive at  Theorem \ref{the:tradeoff_acc_fair}.
\end{proof}
%

\noindent\textbf{\textit{Estimation of Upper Bounds}}. We next describe a methodology for estimating the upper bounds for real-world datasets. 
For Theorem \ref{The:generilized_le_cam} and Theorem \ref{the:tradeoff_acc_fair}, estimating the fraction of the label or sensitive group (i.e., $\alpha$, $\beta$) are quite straightforward. In addition, in order to obtain estimates of these bounds, we need an estimate of $d_{TV}(P_1,P_0), d_{TV}(P_{1}^a,P_{0}^a)$, and $d_{TV}(P_{1}^b,P_{0}^b)$. To this end, we can leverage the fact that TV distance between two distributions is a special case of $F$-divergence, which is known to admit a variational representation \cite{nguyen2010estimating,nguyen2007estimating} expressed as follows:
\begin{align} 
    \hspace{-7pt}D_f(P\parallel Q) =\underset{T(\cdot)}{sup} ~E_{X\sim P}\left[T(X)\right] - E_{X\sim Q}\left[f^{*}(T(X))\right],\label{dp_f_div_variational}
\end{align}
where the function $f^\ast(t)=\underset{x\in dom_f}{sup}\{xt-f(x)\}$ denotes the convex conjugate (also known as the Fenchel conjugate) of the function $f$.
The above variational representation involves a supremum over all possible functions $T(\cdot)$. We can obtain an estimate for TV distance by replacing the supremum over a restricted class of functions. Specifically, if we use a parametric model $T_{\theta}$, (e.g., a neural network) with parameters $\theta$, then taking the supremum over the parameters $\theta$ yields a lower bound on $F$-divergence in \eqref{dp_f_div_variational} as stated next: 
\begin{align} 
    \hspace{-10pt}D_f(P\parallel Q) \hspace{-3pt}
\geq  \underset{\theta}{sup} \hspace{1pt}E_{X\sim P}\left[T_{\theta}(X)\right] - E_{X\sim Q}\left[f^{*}(T_{\theta}(X))\right].\label{dp_lower_bound} 
\end{align}
We use the above variational lower bound to estimate the TV distance as described next. Take the $d_{TV}(P_1, P_0)$ as an example, we need to  estimate TV distance between joint distributions of feature $X$ in different class. The variational lower bound on $F$-divergence in \eqref{dp_lower_bound} can then be estimated as:
\begin{align}\label{dp_div}
    \underset{\theta}{\max} \frac{1}{M}\left(\sum_{m=1}^{M}\hspace{-1pt}T_\theta\left(X_{1}^{(m)}\right) - \sum_{m=1}^{M}\hspace{-1pt}f^\ast\left(T_\theta\left(X_{0}^{(m)}\right)\right)\right), 
\end{align}
where in \eqref{dp_div}, we have replaced the expectation operators with their empirical estimates, and $\{X_{1}^{(m)}\}$ (respectively, $\{X_{0}^{(m)}\}$) denote i.i.d. samples drawn from the distribution $P_{1}$ (respectively, $P_{0}$). The consistency and convergence of the above estimator to the true divergence has been studied in \cite{nguyen2010estimating} under some mild assumptions. For our experiments, we modeled $T_{\theta}$ using two-layer neural networks, each with $10$ hidden nodes and followed by a sigmoid non-linearity activation layer.

\section{Experiments}\label{sec:exp}

In this section, we show the experimental results to verify the tightness of our theoretical bounds. We consider three real world datasets: COMPAS, Adult and Law school admission dataset. We describe the dataset as follows: a) \textit{\underline{COMPAS Dataset}}: This dataset consists of data from $N=7,214$ users ($N_{train} = 5,049$, $N_{test} = 2,165$), with 10 features (including age, prior criminal history, charge degree etc.) which are used for predicting the risk of recidivism in the next two years. 
 b) \textit{\underline{Adult Dataset}}: This dataset includes income related data with 14 features (i.e., age, work class, occupation, education etc.) of $N=45,222$ users ($N_{train} = 32,561$, $N_{test} = 12,661$) to predict whether the income of a person exceeds a threshold (e.g., \$50k) in a year.  c) \textit{\underline{Law School Dataset}}: This dataset includes the admission related data with 7 features (LSAT score, gender, undergraduate GPA etc.) of $N=4,862$ applicants ($N_{train} = 3 ,403$, $N_{test} = 1,459$) to predict the likelihood of passing the bar. For all the above datasets, we use race as the sensitive attribute. Specifically we consider the situation when $|Z|=2, Z \in \{C, O\}$,
 where $C=\text{``Caucasian"}$ or $O=\text{``Other race"}$, corresponding to two groups.

 For the methodology of training a fair classifier, we applied three in-processing mechanisms: (a) Zafar er al. \cite{zafar2017fairness1} ($C_3$ (Covariance)) employ the covariance between sensitive attributes and the signed distance from misclassified data's feature vectors to the classifier decision boundary as a regularization term. (b) Bechavod et al. \cite{bechavod2017learning} ($C_2$ (FNR-FPR)) use the differences in False Negative Rate (FNR) and False Positive Rate (FPR) across subgroups as the regularization term. (c) Zhong et al. \cite{zhong2023learning} ($C_1$ (F-divergence)) propose the F-divergence between the conditional probability of predictions among subgroups as the regularization term.  For the above mechanisms, they added fairness constraints (covariance, FNR-FPR, F-divergence) as a regularization in the loss function to learn a fair classifier subject to equalized odds. 


Fig \ref{fig:upperbound} shows the corresponding tradeoffs achieved by the three fair classifiers as the budget $\epsilon_{EO}$ is increased; we also show the upper bounds of Theorems \ref{The:generilized_le_cam} and \ref{the:tradeoff_acc_fair} as a function of $\epsilon_{EO}$. Notably, our upper bound incorporating EO constraints (Theorem \ref{the:tradeoff_acc_fair}) is tighter compared to the accuracy upper bound without fairness constraints (Theorem \ref{The:generilized_le_cam}). Additionally, it is observed that Theorem \ref{the:tradeoff_acc_fair} closely approximates the upper bound of the classifier's test accuracy under varying EO constraints, until it aligns with Theorem \ref{The:generilized_le_cam}. Our experimental findings reinforce the validity of our theorems, demonstrating a tight correlation between the trade-offs in fairness and accuracy. These results not only align with the trends predicted by our theorems but also underscore the practical applicability of the theoretical framework in diverse real-world scenarios.

\section{Conclusion}

We presented a new upper bound on accuracy for binary classification subject to equalized odds, where the fairness budget is measured by $\epsilon_{EO}$. Our results show that in addition to the fairness budget, relative subgroup sizes $(\beta, 1-\beta)$ as well as the statistical differences across subgroups (measured by $d_{TV}(P_{1}^a,P_{0}^a), d_{TV}(P_{1}^b,P_{0}^b)$) impose a fundamental limit on the accuracy. We also validated these bounds using empirical estimation of TV-distance and compared them with the tradeoffs achieved by various fair classifiers. There are several directions for future work including generalization to other notions of statistical fairness (such as predictive parity); furthermore, it would be interesting to use the upper bounds as a guideline for designing fair classifiers (such as design of group-wise thresholding rules which maximize accuracy).

\section*{Acknowledgment}
This work was supported by NSF grants CCF 2100013, CNS 2209951, CCF 1651492, CNS 2317192, CNS 1822071 and U.S. Department of Energy, Office of Science, Office of Advanced Scientific Computing under Award Number DE-SC-ERKJ422, and by NIH through Award 1R01CA261457-01A1.
\bibliographystyle{ieeetr}
\bibliography{reference}
\end{document}